\documentclass{article}

\usepackage[nonatbib,preprint]{neurips_2025}

\usepackage[utf8]{inputenc} % allow utf-8 input
\usepackage[T1]{fontenc}    % use 8-bit T1 fonts
\usepackage{hyperref}       % hyperlinks
\usepackage{url}            % simple URL typesetting
\usepackage{booktabs}       % professional-quality tables
\usepackage{amsfonts}       % blackboard math symbols
\usepackage{nicefrac}       % compact symbols for 1/2, etc.
\usepackage{microtype}      % microtypography
\usepackage{xcolor}         % colors

\usepackage{graphicx}
\usepackage{amsmath}
\usepackage{amssymb}
\usepackage{multirow}
\usepackage{tabularx}
\usepackage{listings}
\usepackage{sidecap}

\definecolor{codegreen}{rgb}{0,0.6,0}
\definecolor{codegray}{rgb}{0.5,0.5,0.5}
\definecolor{codepurple}{rgb}{0.58,0,0.82}
\definecolor{backcolour}{rgb}{0.95,0.95,0.92}

\lstdefinestyle{mystyle}{
    backgroundcolor=\color{backcolour},   
    commentstyle=\color{codegreen},
    keywordstyle=\color{magenta},
    numberstyle=\tiny\color{codegray},
    stringstyle=\color{codepurple},
    basicstyle=\ttfamily\footnotesize,
    breakatwhitespace=false,         
    breaklines=true,                 
    captionpos=b,                    
    keepspaces=true,                 
    numbers=left,                    
    numbersep=5pt,                  
    showspaces=false,                
    showstringspaces=false,
    showtabs=false,                  
    tabsize=2
}

\newenvironment{teaser}
{\begin{center}
  \begin{minipage}{\textwidth}
}
{\end{minipage}
\end{center}}

\title{Learning Joint ID-Textual Representation for ID-Preserving Image Synthesis}

\author{\textbf{Zichuan Liu} \hspace{6pt} \textbf{Liming Jiang} \hspace{6pt} \textbf{Qing Yan} \hspace{6pt} \textbf{Yumin Jia} \hspace{6pt} \textbf{Hao Kang} \hspace{6pt} \textbf{Xin Lu} \\ \\
  ByteDance Intelligent Creation
}

\begin{document}

\maketitle

\begin{teaser} \label{fig:opening}
\includegraphics[width=0.99\linewidth]{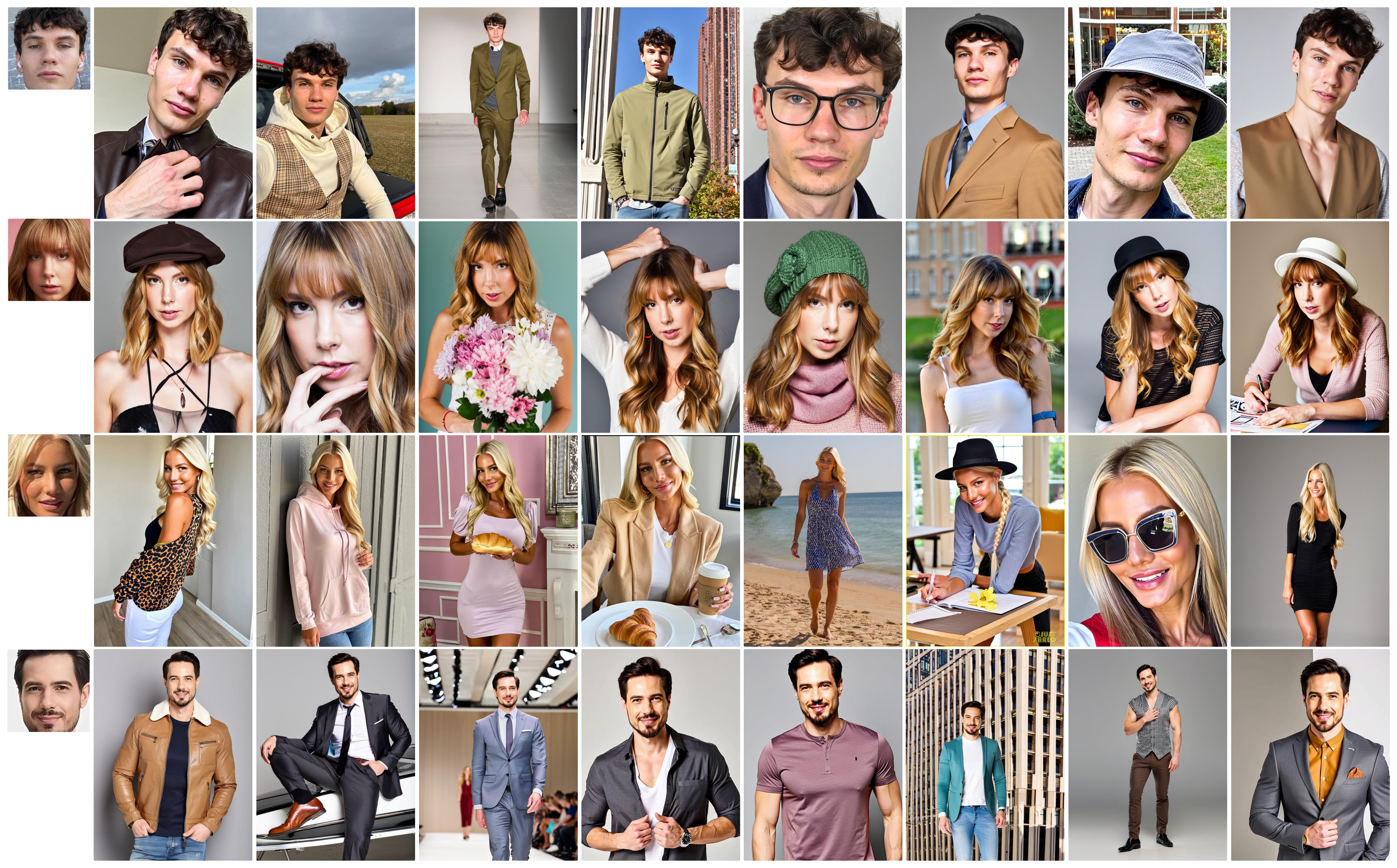}
% \vspace{-0.2in}
{}{FaceCLIP: A novel encoder that learns a joint ID-text representation through multi-modal alignment. Integrating FaceCLIP with SDXL produces exceptional results in ID preservation capability, text alignment, and image quality.}
\end{teaser}

\begin{abstract}
Recent progress in text-to-image (T2I) diffusion models has greatly improved image quality and flexibility. However, a major challenge in personalized generation remains: preserving the subject’s identity (ID) while allowing diverse visual changes. We address this with a new framework for ID-preserving image generation. Instead of relying on adapter modules to inject identity features into pre-trained models, we propose a unified multi-modal encoding strategy that jointly captures identity and text information. Our method, called FaceCLIP, learns a shared embedding space for facial identity and textual semantics. Given a reference face image and a text prompt, FaceCLIP produces a joint representation that guides the generative model to synthesize images consistent with both the subject’s identity and the prompt. To train FaceCLIP, we introduce a multi-modal alignment loss that aligns features across face, text, and image domains. We then integrate FaceCLIP with Stable Diffusion XL, forming a complete synthesis pipeline named FaceCLIP-SDXL. Compared to existing ID-preserving approaches, our method produces more photorealistic portraits with better identity retention and text alignment. Extensive experiments demonstrate that FaceCLIP-SDXL outperforms prior methods in both qualitative and quantitative evaluations.
\end{abstract}
\section{Introduction}
\noindent
Recent advancements in text-to-image (T2I) diffusion models~\cite{nichol2021glide,ramesh2022hierarchical,rombach2022high,saharia2022photorealistic} have substantially improved the quality and flexibility of image generation. A central objective of this progress is personalized generation, which aims to preserve the subject’s identity (ID) from reference images while allowing versatile modifications of other visual attributes.

\par\noindent
ID-preserving image synthesis has undergone a significant evolution, transitioning from tuning-based approaches to tuning-free methods. Tuning-based methods~\cite{ruiz2023dreambooth,wei2023elite,hu2022lora,gal2022image} achieve ID preservation through test-time fine-tuning. These approaches adapt a pre-trained generative model using a small number of reference images, biasing the model to generate images that resemble the reference subject. Although such methods can offer a certain level of identity preservation, they suffer from limited semantic controllability, high computational cost, and poor real-time performance. In contrast, tuning-free methods~\cite{ye2023ip,shi2024instantbooth,guo2025pulid,wang2024instantid,he2024uniportrait,sara2025ipcompose,qian2024omni,han2024face} provide zero-shot personalization solutions by integrating identity features into pre-trained foundation models, such as Stable Diffusion XL (SDXL)~\cite{podell2023sdxl} and FLUX~\cite{guo2025pulid}. These approaches typically introduce parameterized plugin modules or adapters that adapt identity features (e.g., face embeddings) and inject them into the generation process. The adapters guide the foundation model to produce identity-consistent images without requiring test-time fine-tuning. The effectiveness of this ID preservation paradigm has been demonstrated in recent works such as InstantID~\cite{wang2024instantid} and PuLID~\cite{guo2025pulid}, which generate high-quality images with satisfactory identity consistency and semantic alignment with textual prompts. However, since these adapter modules are built upon and trained alongside a fixed pre-trained foundation model, injecting identity features may interfere with the model’s original generation capabilities. This often results in reduced diversity and perceptually unnatural images.
\par
\noindent
In this paper, we address these limitations from a new perspective. Inspired by Arc2Face~\cite{papantoniou2024arc2face}, we formulate portrait image generation as a forward probabilistic process conditioned on both identity and textual semantics. ID-preserving generation is thus realized by sampling from this conditional distribution, which can be learned by a diffusion model with appropriate conditioning inputs. To obtain effective conditioning signals, we propose a novel multi-modal encoder, \emph{FaceCLIP}, which jointly encodes identity features and text semantics into a unified representation. This representation preserves discriminative information from both modalities and serves as the prior condition for portrait image generation. To train FaceCLIP, we develop a multi-modal alignment algorithm that optimizes a joint embedding space using a multi-modal alignment loss. This loss explicitly aligns the joint representation with the original face embedding space, the text embedding space, and the image embedding space.
\par
\noindent
Furthermore, we integrate FaceCLIP into Stable Diffusion XL (SDXL)~\cite{podell2023sdxl} and build an ID-preserving foundation model, \emph{FaceCLIP-SDXL}. Unlike existing plugin-based ID preservation approaches, our goal is to integrate identity preservation capability directly into the foundation model. This design fully leverages the capacity of the foundation model and the diversity of large-scale data. Through joint encoding of identity and text, our method promotes deeper interaction between visual identity features and textual semantics during diffusion model training. As a result, FaceCLIP-SDXL generates photorealistic images with consistent subject identity and accurate semantic alignment.
\par
\noindent
Extensive experiments demonstrate that, when trained with the proposed alignment loss, FaceCLIP effectively captures both identity and semantic information. The resulting ID-preserving foundation model outperforms existing methods both quantitatively and qualitatively in terms of identity preservation and text alignment. Moreover, compared to previous approaches, FaceCLIP-SDXL produces more diverse and photorealistic images with exceptional detail in skin texture and body structure.
The contributions of this paper are summarized as follows:
\begin{itemize}
    \item We revisit existing reconstruction-based ID-preserving image generation methods and reformulate the task as a sampling process from a forward distribution conditioned on identity and text semantics.
    \item We propose a novel multi-modal encoder, FaceCLIP, that encodes both identity and textual information into a unified representation, serving as a prior for downstream generation tasks.
    \item We design a multi-modal alignment algorithm for training FaceCLIP, enabling it to learn discriminative joint representations from both identity and textual modalities.
    \item We demonstrate the effectiveness of FaceCLIP for ID-preserving image generation by integrating it with SDXL. Experimental results show that the proposed FaceCLIP-SDXL achieves state-of-the-art performance in identity preservation, text adherence, and image fidelity.
\end{itemize}

\section{Related Works}

\subsection{Multimodal Alignment and Fusion}
\noindent
Integrating information from multiple modalities can significantly enhance the performance of machine learning models. Recent progress in multi-modal learning has shown promising results across various applications, including image captioning, video summarization, machine translation, and image generation~\cite{gabeur2020multi,fei2024video,zhu2023minigpt,lin2023video,li2023llava,li2024llava,bai2025qwen2,akhmerov2019research}. However, two key challenges remain in effectively utilizing multi-modal data: alignment and fusion. Alignment~\cite{baltruvsaitis2018multimodal,li2021align} aims to establish semantic correspondence across different modalities, ensuring that representations from diverse sources are mapped into a shared latent space. This enables the model to correlate and reason jointly over multi-modal inputs. Fusion~\cite{barua2023systematic,tian2019multimodal,shankar2022progressive,snoek2005early}, on the other hand, focuses on combining multiple modalities into a unified representation, leveraging their complementary strengths to improve model performance. Our approach incorporates both alignment and fusion. To achieve a unified representation for identity preservation, we integrate identity and textual information through fusion. Furthermore, we re-align this joint representation with the original embedding spaces to retain the distinctive characteristics of both modalities. This dual process ensures effective identity preservation while maintaining semantic coherence between input modalities.

\subsection{Customized Image Synthesis}
\noindent
Customized image synthesis aims to generate images of a real-world object or person under diverse contexts. Existing approaches can be broadly categorized into tuning-based and tuning-free methods. Tuning-based methods fine-tune pre-trained generative models~\cite{nichol2021glide,ramesh2022hierarchical,rombach2022high,saharia2022photorealistic,esser2024scaling,tong2023improving,ma2024sit,liu2022flow} using a limited number of reference images of the target object or individual. These methods bias the generative model toward a specific identity while preserving a certain degree of editability. However, they require test-time optimization for each subject, making them computationally expensive and unsuitable for real-time or scalable applications. Tuning-free methods~\cite{ye2023ip,shi2024instantbooth,guo2025pulid,wang2024instantid,he2024uniportrait,sara2025ipcompose,qian2024omni,han2024face} enable subject-driven image generation by leveraging identity features extracted from reference images. Typically, these methods use a learnable adapter module to inject identity features into a pre-trained generative model. The extracted features are projected into token representations and blended with text tokens via a cross-attention mechanism~\cite{lin2022cat}. Various works have demonstrated the effectiveness of tuning-free methods. For example, AnyDoor~\cite{chen2024anydoor} utilizes DINOv2 features~\cite{oquab2023dinov2} for subject-preserving background replacement, while IMPRINT~\cite{song2024imprint} proposes a context-agnostic ID-preserving training strategy to enhance appearance retention. Other methods such as IP-Adapter, InstantID, PuLID, and Arc2Face~\cite{ye2023ip,wang2024instantid,guo2025pulid,papantoniou2024arc2face} adopt ArcFace embeddings~\cite{deng2019arcface} to achieve high-quality portrait synthesis.
\par
\noindent
Our method also belongs to the tuning-free category. However, instead of injecting identity features into intermediate layers of the base model via an adapter, we propose training an encoder that produces a unified prior representation from different modalities to guide the generation process. Unlike previous approaches that formulate ID preservation as a reconstruction task, FaceCLIP-SDXL learns a forward distribution over portrait images conditioned on identity and textual semantics. The model is trained on large-scale datasets annotated with both identity and semantic information. Our method is inspired by the recently proposed face foundation model, Arc2Face~\cite{papantoniou2024arc2face}, which trains a diffusion model for face synthesis based on Stable Diffusion 1.5 (SD1.5)~\cite{rombach2022high}. In Arc2Face, face embeddings are treated as special tokens processed by a trainable CLIP-based text encoder. The generative model and text encoder are jointly optimized to synthesize diverse face images conditioned on a reference image. While our method shares similar idea of modeling ID-preserving generation as a conditional sampling process, it extends this formulation to incorporate both identity and global semantic attributes. In contrast to Arc2Face, which is limited to generating headshot images, our FaceCLIP-SDXL is capable of generating full portrait images with diverse perspectives and camera distances.

\section{Methodology}

\subsection{Problem Formulation}
\noindent
We formulate the data distribution of portrait images as a conditional distribution based on the subject's identity and textual description, denoted as $p(x\mid e)$. Here, $e$ represents a unified representation of both identity and text conditions, defined as:
\begin{equation} \label{eq:encoder}
e = \mathcal{H}_{\theta}(c_{\text{t}}, c_{\text{r}}),
\end{equation}
where $\mathcal{H}_{\theta}$ denotes the multi-modal FaceCLIP encoder, and $c_{t}$ and $c_{r}$ denote the input text prompt and a reference image indicating identity information, respectively. We employ a diffusion model $\epsilon_{\phi}$ to approximate $p(x \mid e)$, and the training objective is formulated as:
\begin{equation}
L_{\text{DM}} = \mathbb{E}_{x_0, e, \epsilon \sim \mathcal{N}(\mathbf{0}, \mathbf{I}), t} \left[ \left\| \epsilon - \epsilon_{\phi}(x_t, t, e) \right\|^2_2 \right],
\end{equation}
where $x_0$ is a data sample drawn from $p(x \mid e)$, $\epsilon$ is Gaussian noise sampled from a standard normal distribution, and $x_t$ denotes the noisy latent at time step $t$. Given a dataset consisting of tuples $(x_0, c_{\text{t}}, c_{\text{r}})$, our objective is twofold: (1) to learn an encoder $\mathcal{H}_{\theta}$ that jointly embeds identity and textual information into a unified representation $c$, and (2) to train a diffusion model $\epsilon_{\theta}$ to approximate the conditional data distribution $p(x \mid e)$, guided by this unified prior.

\subsection{Learning Joint Identity-text Encoding} 
\noindent
The existing text-to-image (T2I) generative models rely on text embeddings as semantic priors to guide the generation process. To enable ID-preserving generation, we treat identity information as a distinct semantic modality and integrate it with text to jointly guide the generation process. These two modalities are fused into a unified representation \( e \in \mathbb{R}^{b \times L \times d_c} \) via the transformation defined in Equation~\ref{eq:encoder}, where \( b \), \( L \), and \( d_c \) denote batch size, sequence length, and feature dimension, respectively.
Inspired by CLIP~\cite{radford2021learning}, we develop a standalone pre-training algorithm to learn the parameters \( \theta \), ensuring that the joint representation \( e \) effectively preserves both identity and textual semantics. Similar to how CLIP aligns image and text embeddings, we align the joint representation \( e \) with the image embedding space \( e_I \in \mathbb{R}^{b \times d_t} \) obtained from a pre-trained CLIP visual encoder using a contrastive loss. The mapping from an image to its corresponding embedding is given by
$e_I = \mathcal{F}_{\text{im}}(x_0)$,
where \( \mathcal{F}_{\text{im}} \) denotes the CLIP image encoder and \( x_0 \) is the input image.
To further enhance identity preservation, we introduce an additional contrastive loss that aligns \( e \) with the identity embedding space \( e_{r_{\text{cls}}} \in \mathbb{R}^{b \times d_r} \), derived from a face recognition backbone such as ArcFace~\cite{deng2019arcface}. The face embedding is obtained via
$e_{r_{\text{cls}}} = \mathcal{F}_{\text{id}}(c_r)$,
where \( \mathcal{F}_{\text{id}} \) is the face encoder and \( c_r \) is an aligned reference face image.
To accelerate convergence and improve semantic expressiveness, we also align \( e \) with a text embedding space \( e_{t_{\text{cls}}} \in \mathbb{R}^{b \times d_t} \), obtained from a pre-trained CLIP text encoder:
$e_{t_{\text{cls}}} = \mathcal{F}_t(c_t)$,
where \( \mathcal{F}_t \) is the text encoder and \( c_t \) denotes the input text prompt.
The overall pre-training objective is a sum of three contrastive losses:
\begin{equation} \label{eq:loss}
\mathcal{L} = \mathcal{L}_c(e_{c \to t}, e_I) + \mathcal{L}_c(e_{c \to r}, e_{r_{\text{cls}}}) + \mathcal{L}_c(e_{c \to t}, e_{t_{\text{cls}}}),
\end{equation}
where \( e_{c \to t} \in \mathbb{R}^{b \times d_t} \) and \( e_{c \to r} \in \mathbb{R}^{b \times d_r} \) are linear projections of the joint embedding \( e \), used to match the dimensionality of image and ID embedding spaces, respectively.
\( \mathcal{L}_c(\cdot, \cdot) \) is the contrastive loss function proposed in \cite{radford2021learning}.
% \begin{lstlisting}[language=Python]
% # x1 [b, d]: projected embedding batch
% # x2 [b, d]: projected embedding batch
% import numpy as np

% def contrastive_loss(x1, x2):
%     logits = np.dot(x1, x2.T)
%     loss1 = cross_entropy_loss(logits, axis=0)
%     loss2 = cross_entropy_loss(logits, axis=1)
%     loss = (loss1 + loss2) / 2
%     return loss
% \end{lstlisting}

\subsection{Pre-training Workflow}
\noindent
The high-level architecture of our FaceCLIP encoder $\mathcal{H}$ and training workflow is illustrated in Figure~\ref{fig:system} (a). The FaceCLIP encoder processes a batch of aligned face images \( c_r \in \mathbb{R}^{b\times h_r\times w_r\times d_r} \) and text prompts \( c_t \) as input, outputting fused embeddings \( e \). Here, \(h_{<*>}\) and \(w_{<*>}\) denotes the spatial size of an image. Within the encoder, the face image is processed by a face encoder into a face class embedding \( e_{r_{\text{cls}}} \in \mathbb{R}^{b\times d_{r_{\text{cls}}}} \) and face patch features \( e_{r_{\text{pat}}} \in \mathbb{R}^{b\times L_r\times d} \). Meanwhile, the text prompt is converted into text class embeddings \( e_{t_{\text{cls}}} \in \mathbb{R}^{d_{t_{\text{cls}}}} \) and text patch embeddings \( e_{t_{\text{pat}}} \in \mathbb{R}^{b\times L\times d} \). The fused representation \( e \) is obtained by feeding \( e_{r_{\text{pat}}} \) and \( e_{t_{\text{pat}}} \) into the Fusion Module. As defined in the previous section, \( e \) serves as a unified representation of both identity and textual information. To enable the FaceCLIP encoder to learn a joint identity-text embedding space, we extend the vision-language pretraining method proposed in~\cite{radford2021learning} with an additional identity alignment loss. As depicted in Figure~\ref{fig:system} (a), \( e \) is further projected into projected face embeddings \( e_{c\to r} \) and projected text embeddings \( e_{c\to t} \). Similar to CLIP, we align \( e_{c\to t} \) with image embeddings \( e_I \) extracted from target images \( x_0 \) by a CLIP vision encoder using a contrastive loss~\cite{chen2020simple}. However, applying contrastive loss solely on \( e_{c\to t} \) and \( e_I \) may lead to trivial text-image alignment. To address this, we introduce an additional contrastive loss to explicitly align \( e_{c\to r} \) with \( e_{r_{\text{cls}}} \), ensuring that \( e \) preserves identity information. During training, the face encoder, CLIP text encoder, CLIP vision encoder, and the image projection layer are freeze. 
 
\subsection{Fusion Module} 
\begin{figure}
\centering
\includegraphics[width=\textwidth]{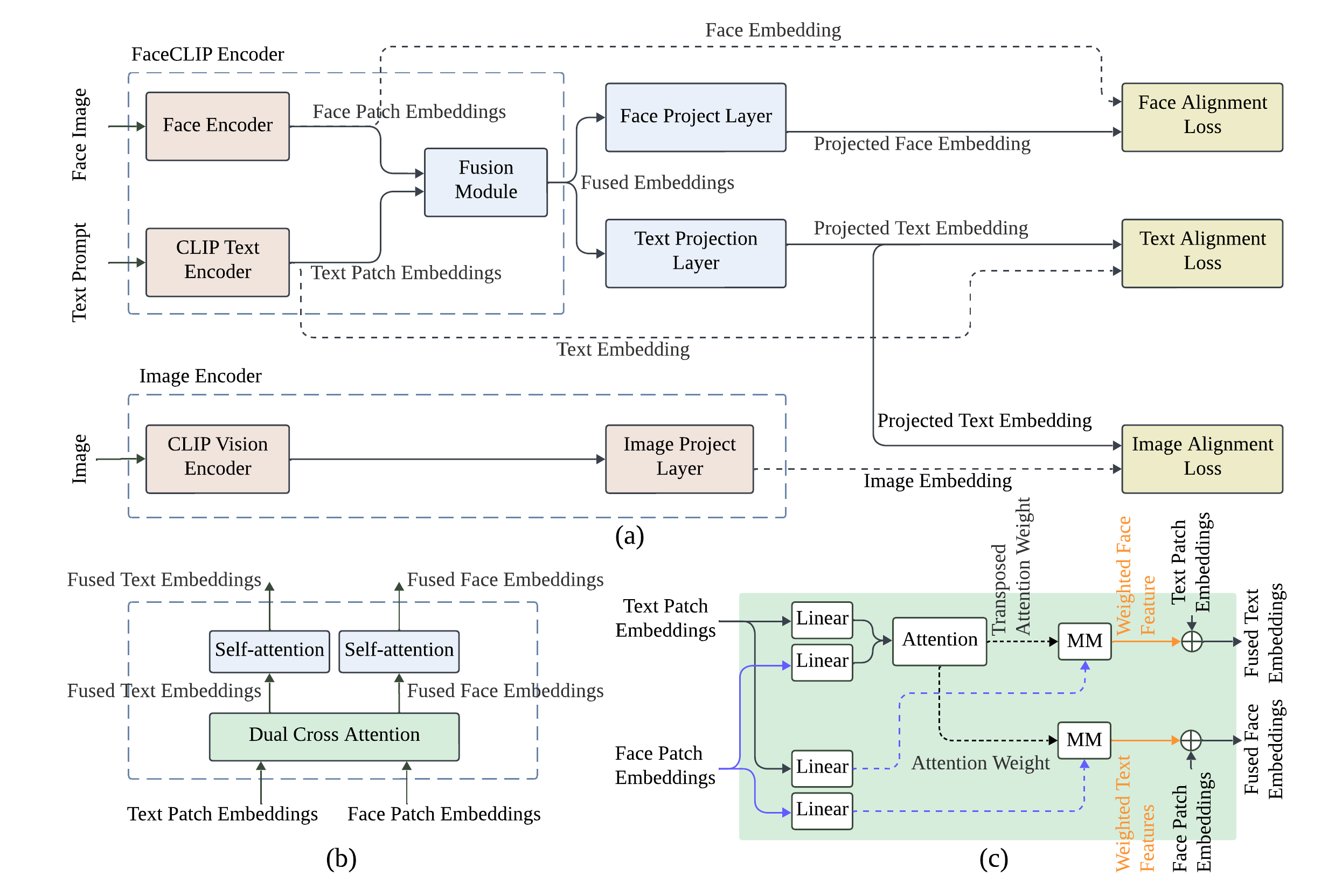}
% \vspace{-0.1in}
\caption{(a) FaceCLIP architecture and pre-training workflow.The modules labeled in red are freezed and the modules labeled in blue are unfreezed during pre-training; (b) Architecture of Feature Fusion Block in Fusion Module; (c) Detailed computation graph of Dual Cross-Attention.}
\label{fig:system}
\vspace{-0.4cm}
\end{figure}

\noindent
The Fusion Module (FM) is the key component responsible for integrating identity features and text features. It consists of multiple cascaded Feature Fusion (FF) blocks, with the architecture of a single FF block depicted in Figure~\ref{fig:system} (b) and (c). Each FF block takes text patch embeddings \( e_{t_{\text{pat}}} \) and face patch embeddings \( e_{r_{\text{pat}}} \) as input and outputs fused text embeddings \( \hat{e}_{t_{\text{pat}}} \) and fused face embeddings \( \hat{e}_{r_{\text{pat}}} \). Within each FF block, \( e_{t_{\text{pat}}} \) and \( e_{r_{\text{pat}}} \) first pass through a Dual Cross-Attention (DCA) module, which integrates text and identity information using a cross-attention mechanism. The mixed embeddings are then processed by two separate self-attention (SA) layers, yielding the final fused text embeddings \( \hat{e}_{t_{\text{pat}}} \) and fused face embeddings \( \hat{e}_{r_{\text{pat}}} \). Finally, the fused text embeddings from the last FF block serve as the joint identity-text representation \( e \).
% \begin{figure}
% \centering
% \includegraphics[width=\textwidth]{figures/system.pdf}
% % \vspace{-0.1in}
% \caption{FaceCLIP architecture and pre-training workflow. The modules labeled in red are freezed and the modules labeled in blue are unfreezed during pre-training.}
% \label{fig:system}
% % \vspace{-0.4cm}
% \end{figure}

\subsection{Regularizing Pre-training with Image-text Data}
\noindent
Since the volume of main training data (main dataset) structured into tuples \( (x_0, c_t, c_r) \) is relatively small compared with the dataset to train the CLIP model, solely using these training data can degrade the performance of text alignment. Thus, we apply additional an internal large-scale image-text dataset, as the guided dataset to preserve the text alignment capability of our model. During pre-training, we randomly replace the data from main dataset with the data from guided dataset with probability of $\lambda$. The FaceCLIP encoder processes the data from guided dataset by zero-out the input of the face encoder. Experiments of zero-shot classification show that this approach can effectively preserve FaceCLIP's zero-shot classification capability, which directly related to its text alignment performance.

\subsection{Training Diffusion Model with Joint Identity-text Priors}
\noindent
We adopt SDXL~\cite{podell2023sdxl} as the base diffusion model to approximate the data distribution \( p(x \mid e) \), where \( e \) denotes the joint identity and text condition. Since the original SDXL framework relies on two text encoders to provide semantic priors, we independently train two FaceCLIP encoders to replace the native OpenAI-CLIP-L-14 and OpenCLIP-bigG-14 encoders. Specifically, we implement the text encoder within the FaceCLIP architecture using the OpenAI-CLIP-L-14 and OpenCLIP-bigG-14 backbones, respectively. The resulting multi-modal encoders are referred to as \emph{FaceCLIP-L/14} and \emph{FaceCLIP-bigG-14}. Given training data tuples \( (x_0, c_t, c_r) \), we generate noisy latents \( x_t \) following the standard DDPM formulation:
\begin{equation}
    x_t = \sqrt{\alpha} \cdot x_0 + \sqrt{1 - \alpha} \cdot \epsilon,
\end{equation}
where \( \alpha \) denotes the noise scheduling parameter and \( \epsilon \) is sampled from a standard normal distribution. The diffusion model \( \epsilon_{\phi} \) is then optimized according to Equation~\ref{eq:loss}.
\par
\noindent
In contrast to prior adapter-based approaches, which train only a small adapter module while keeping the foundation model fixed, our method fully adapts the entire diffusion model to the joint identity-text condition. In existing methods, content outside the facial region is predominantly controlled by the pre-trained foundation model, which is often trained on non-domain-specific data. As a result, the adapter is only capable of blending the identity information into localized regions, with limited influence on the overall layout and visual style. By contrast, our approach leverages the full modeling capacity of the foundation model and encourages richer interaction between identity and text features. The unified embedding space produced by FaceCLIP enables the diffusion model to capture complex relationships between the subject’s facial appearance and the surrounding visual context. This leads to photorealistic image generation with natural subject rendering and coherent global composition.

% \begin{figure}
% \centering
% \includegraphics[width=0.7\textwidth]{figures/fusion.pdf}
% % \vspace{-0.15in}
% \caption{Fusion Module: (a) Architecture of Feature Fusion Block in Fusion Module; (b) Detailed computation graph of Dual Cross-Attention.}
% \label{fig:fusion_module}
% % \vspace{-0.2in}
% \end{figure}
\section{Experiments}
\noindent
We conduct a series of experiments to evaluate the proposed method from multiple perspectives: (1) to validate that the FaceCLIP encoder learns distinctive representations for both identity and textual semantics; (2) to assess the effectiveness of the FaceCLIP representation and pre-training strategy in enabling ID-preserving image generation; (3) to evaluate the overall performance of FaceCLIP-SDXL by comparing it with existing ID preservation. Specifically, the text alignment performance is assessed via zero-shot classification accuracy on ImageNet-1k~\cite{gal2022image}. The quality of identity alignment is examined by visualizing the fused embedding \( e \) using t-SNE~\cite{cai2022theoretical}. Additionally, we conduct an ablation study to investigate the impact of the FaceCLIP representation on image generation quality. Finally, we compare our method against existing ID preservation approaches through both quantitative evaluations and qualitative assessments.

\subsection{Setting}
\noindent
\textbf{Implementation Details.} 
We implement a FaceCLIP-L-14, a FaceCLIP-bigG-14, and FaceCLIP-SDXL for all experiments using PyTorch. All models are trained in a distributed manner using PyTorch's built-in Distributed Data Parallel (DDP) library with numerical precision set to $bf16$. FaceCLIP-L-14 and FaceCLIP-bigG-14 are trained on 64 NVIDIA H100 GPUs with batch sizes of 24K and 16K, respectively. FaceCLIP-SDXL is trained on 128 NVIDIA H100 GPUs with a total batch size of 4096 and 512. To mitigate weight stagnation caused by $bf16$, we adopt the AnyPrecisionAdamW optimizer \cite{anyprecision}, setting $\beta$ to $(0.9, 0.999)$ and weight decay to 0.01. The learning rate for all models is configured as $2e-5$. 

\noindent
\textbf{Datasets.} We construct a large-scale Single-Per-Single-View (SPSV) dataset comprising tuples of images, text descriptions, and reference images to train FaceCLIP encoder and FaceCLIP-SDXL. The reference images are aligned face crops with a spatial size of $448\times 448$. Our dataset includes nine public datasets \cite{cao2018vggface2,guo2016ms,li2020celeb,zhu2022celebv,karras2019style,xie2022vfhq,kapitanov2023easyportrait,yu2023celebv,li2024cosmicman} alongside internal datasets. After rigorous preprocessing and filtering, we obtain a total of 43 million data samples with portrait images and captions annotated by InternVL \cite{chen2024internvl}. To preserve text alignment capability, we use an internal large-scale image-text dataset as guided data. We evaluate image synthesis performance on the Internal-v1 validation set and Unsplash-50 \cite{guo2025pulid}. Internal-v1 includes 15 identities, each associated with 200 text prompts describing variations in location, weather, lighting conditions, and subject behavior. Unsplash-50 consists of 50 high-resolution portrait images with corresponding descriptions. 

\noindent
\textbf{Baselines.} We evaluate our method in tasks of face synthesis and ID-preserved generation. In face synthesis, we select state-of-art Arc2Face as the baseline. In ID-preserved generation, while advanced methods such as PuLID-FLUX \cite{guo2025pulid} and FLIX.1-dev IP-Adapter \cite{flux-ipa} produce high-quality images, their extensive computational demands make them impractical for rapid validation. To ensure a fair comparison, we select InstantID \cite{wang2024instantid} and PuLID-SDXL \cite{guo2025pulid} as baselines. Both methods utilize the same base model as FaceCLIP-SDXL, making them suitable benchmarks for evaluation. 

\noindent
\textbf{Evaluation.} We evaluate ID-preserving image synthesis using three key metrics: Face Similarity \cite{deng2019arcface}, CLIP Score \cite{radford2021learning}, and Fr\'echet Inception Distance (FID) \cite{heusel2017gans}. These metrics assess identity similarity, text alignment, and perceptual quality, respectively. Face Similarity is measured as the cosine similarity between the face embeddings of the reference and generated images. CLIP Score is computed as the cosine similarity between the generated image's embedding and the corresponding text prompt's embedding. FID is calculated between the generated images and a reference set of over 1400 images, evaluating the distributional distance between real and generated samples. Additionally, we perform human evaluation as a complementary assessment to capture qualitative aspects that automated metrics may not fully reflect. We conduct two user studies to validate the effectiveness of face synthesis and ID preserved generation. We recruited 17 participants from diverse backgrounds, e.g. professionals, researchers, engineers, designers, to reduce bias. 

\subsection{Text Alignment Verification}
\noindent
We evaluate zero-shot classification accuracy on ImageNet-1K \cite{deng2009imagenet} of FaceCLIP, following the official zero-shot classification protocol \cite{radford2021learning}. Specifically, visual embeddings are extracted via the CLIP vision encoder, while classifier weights are obtained by feeding 1000 class-related texts into the FaceCLIP encoder using a black face image. The results, presented in Table \ref{tab:zero_shot}, show that FaceCLIP achieves comparable zero-shot classification performance to the baseline OpenCLIP model. FaceCLIP-L-14 slightly outperforms OpenCLIP, achieving top-1/top-5 accuracy of 75.3/94.9, while FaceCLIP-bigG-14 lags slightly behind its OpenCLIP counterpart with 76.9/95.1. Due to computational constraints, FaceCLIP-bigG-14 was trained with a batch size of 16K, significantly smaller than the official implementation, indicating that performance could improve with increased batch size. As shown in Table \ref{tab:zero_shot}, FaceCLIP encoders trained with guided data exhibit a notable performance drop. Therefore, it is important to incorporate guided datasets for maintaining text alignment performance. 

% \begin{table}
% \small
% \centering
% \begin{tabularx}{0.3\textwidth}{lcc}
% \toprule
% Model & Top1 / Top5 \\
% \midrule
% OpenCLIP-L-14 & 75.2 / 94.3 \\
% OpenCLIP-bigG-14 &  80.1 / 96.0  \\ \hline
% FaceCLIP-L-14$^\prime$ & 45.8 / 64.2 \\
% FaceCLIP-bigG-14$^\prime$ & 47.2 / 69.9  \\ \hline
% FaceCLIP-L-14 & 75.3 / 94.9 \\
% FaceCLIP-bigG-14 & 76.9 / 95.1  \\
% \bottomrule
% \end{tabularx}
% \caption{Top-1 and Top-5 classification accuracy of FaceCLIP-L-14 and FaceCLIP-bigG-14 on ImageNet-1K. The symbol ($^\prime$) denotes a FaceCLIP encoder pre-trained without guided data. In this case, the zero-shot classification accuracy drops significantly, indicating that guided data is crucial for preserving text alignment capability.}
% \label{tab:zero_shot}
% \vspace{-0.4cm}
% \end{table}

% \begin{table}
% \small
% \centering
% \begin{tabularx}{0.4\textwidth}{lcc}
% \toprule
% Model & Top1 / Top5 \\
% \midrule
% OpenCLIP-L-14 & 75.2 / 94.3 \\
% OpenCLIP-bigG-14 &  80.1 / 96.0  \\ \hline
% FaceCLIP-L-14$^\prime$ & 45.8 / 64.2 \\
% FaceCLIP-bigG-14$^\prime$ & 47.2 / 69.9  \\ \hline
% FaceCLIP-L-14 & 75.3 / 94.9 \\
% FaceCLIP-bigG-14 & 76.9 / 95.1  \\
% \bottomrule
% \end{tabularx}
% \vspace{0.2cm}
% \caption{Top-1 and Top-5 classification accuracy of FaceCLIP-L-14 and FaceCLIP-bigG-14 on ImageNet-1K. The symbol ($^\prime$) denotes a FaceCLIP encoder pre-trained without guided data.}
% \label{tab:zero_shot}
% % \vspace{-0.4cm}
% \end{table}

\begin{table}
\small
\centering
\begin{tabularx}{\textwidth}{lc|lc|lc}
\toprule
Model & Top1/5 & Model & Top1/5 & Model & Top1/5  \\
\midrule
OpenCLIP-L-14 & 75.2 / 94.3  & FaceCLIP-L-14$^\prime$ & 45.8 / 64.2 & FaceCLIP-L-14 & 75.3 / 94.9\\
OpenCLIP-bigG-14 &  80.1 / 96.0  & FaceCLIP-bigG-14$^\prime$ & 47.2 / 69.9 & FaceCLIP-bigG-14 & 76.9 / 95.1\\ \hline
\bottomrule
\end{tabularx}
\vspace{0.1cm}
\caption{Top-1 and Top-5 classification accuracy of FaceCLIP-L-14 and FaceCLIP-bigG-14 on ImageNet-1K. The symbol ($^\prime$) denotes a FaceCLIP encoder pre-trained without guided data.}
\label{tab:zero_shot}
\vspace{-0.4cm}
\end{table}

\subsection{Identity Alignment Verification} 
\begin{figure}
\centering
\includegraphics[width=\textwidth]{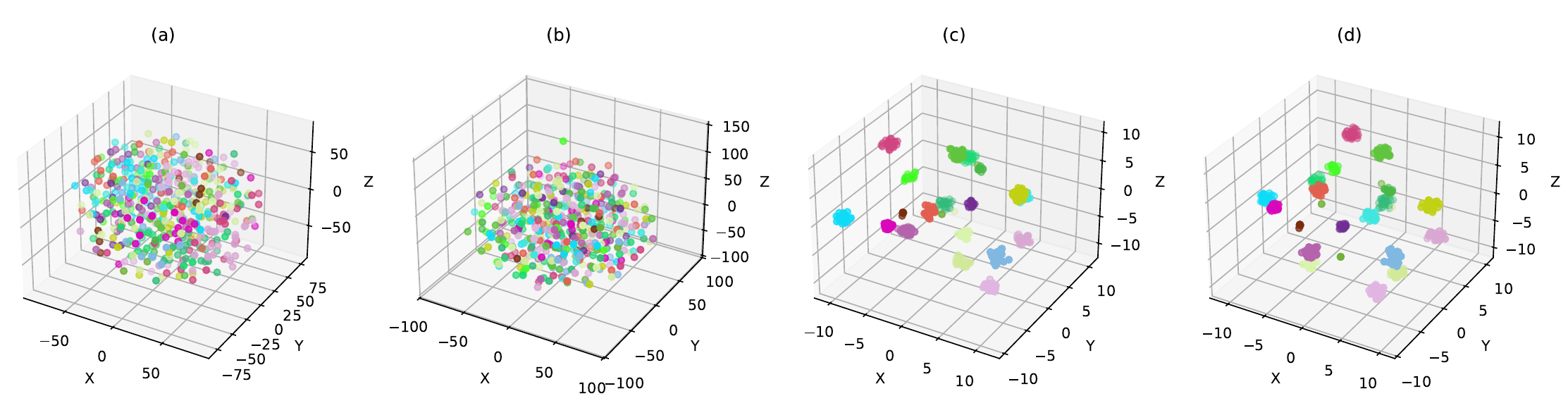}
% \vspace{-0.1in}
\caption{Visualization of subspaces learned by FaceCLIP-L-14 and FaceCLIP-bigG-14. We visualize 500 aligned face images across 20 identities. (a) and (b) depict subspaces learned via diffusion loss without pre-training, whereas (c) and (d) illustrate subspaces from pre-trained FaceCLIP-L-14 and FaceCLIP-bigG-14, respectively.}
\label{fig:tsne}
\vspace{-0.4cm}
\end{figure}

\noindent
We verify ID alignment effectiveness by visualizing fused embeddings from 20 identities using t-SNE \cite{van2009learning}. To isolate the impact of pre-training, we compare the subspaces learned by FaceCLIP encoders pre-trained using our workflow versus those trained directly with diffusion loss. As illustrated in Figure \ref{fig:tsne} (a) and (b), the subspaces learned via diffusion loss alone lack meaningful identity-related structures. Conversely, as seen in Figure \ref{fig:tsne} (c) and (d), our pre-trained FaceCLIP encoders encode clear, discriminative clusters corresponding to distinct identities. This suggests that direct diffusion training does not provide sufficient identity representation, confirming the necessity of our pre-training approach for ID preservation in downstream synthesis tasks. 

\subsection{Identity-Preserved T2I}
% \noindent \vspace{-0.4in}
% \subsubsection{Face Synthesis}
% \noindent
% \textbf{Quantitative Assessment.} To ensure a fair comparison with Arc2Face, which is developed based on Stable Diffusion 1.5 (SD1.5), we compare our downgraded variant, FaceCLIP-SD1.5, against Arc2Face in terms of face similarity and image fidelity, using both quantitative and qualitative evaluations. As shown in Table~\ref{tab:arc2face}, FaceCLIP-SD1.5 outperforms Arc2Face, achieving higher face similarity scores of 0.930 on Internal-v1 and 0.919 on Unsplash-50. Moreover, FaceCLIP-SD1.5 demonstrates superior image quality, reflected by lower FID scores of 5.823 and 6.012, respectively. Qualitative comparisons are illustrated in Figure~\ref{fig:arc2face}, where FaceCLIP-SD1.5 generates images with enhanced realism, improved fidelity, and more natural tonal rendering, while maintaining strong identity consistency.
% \input{tables/arc2face}
% \noindent
% \textbf{User Study.} We further conducted a user study to compare FaceCLIP-SD1.5 with Arc2Face, focusing on face similarity and image fidelity. The selection rate in favor of FaceCLIP-SD1.5 reached 74.2\%, compared to 25.8\% for Arc2Face, indicating a clear human preference for our method in terms of perceived quality and identity resemblance.

% \begin{figure}[h]
% \centering
% \includegraphics[width=0.9\textwidth]{figures/arc2face_2.pdf}
% \vspace{-0.15in}
% \caption{Face images by Arc2Face and FaceCLIP-SD1.5.}
% \label{fig:arc2face}
% \vspace{-0.2in}
% \end{figure}

% \vspace{-0.1in}
\subsubsection{ID Preservation Comparison}
\begin{figure} 
\centering
\includegraphics[width=0.95\textwidth]{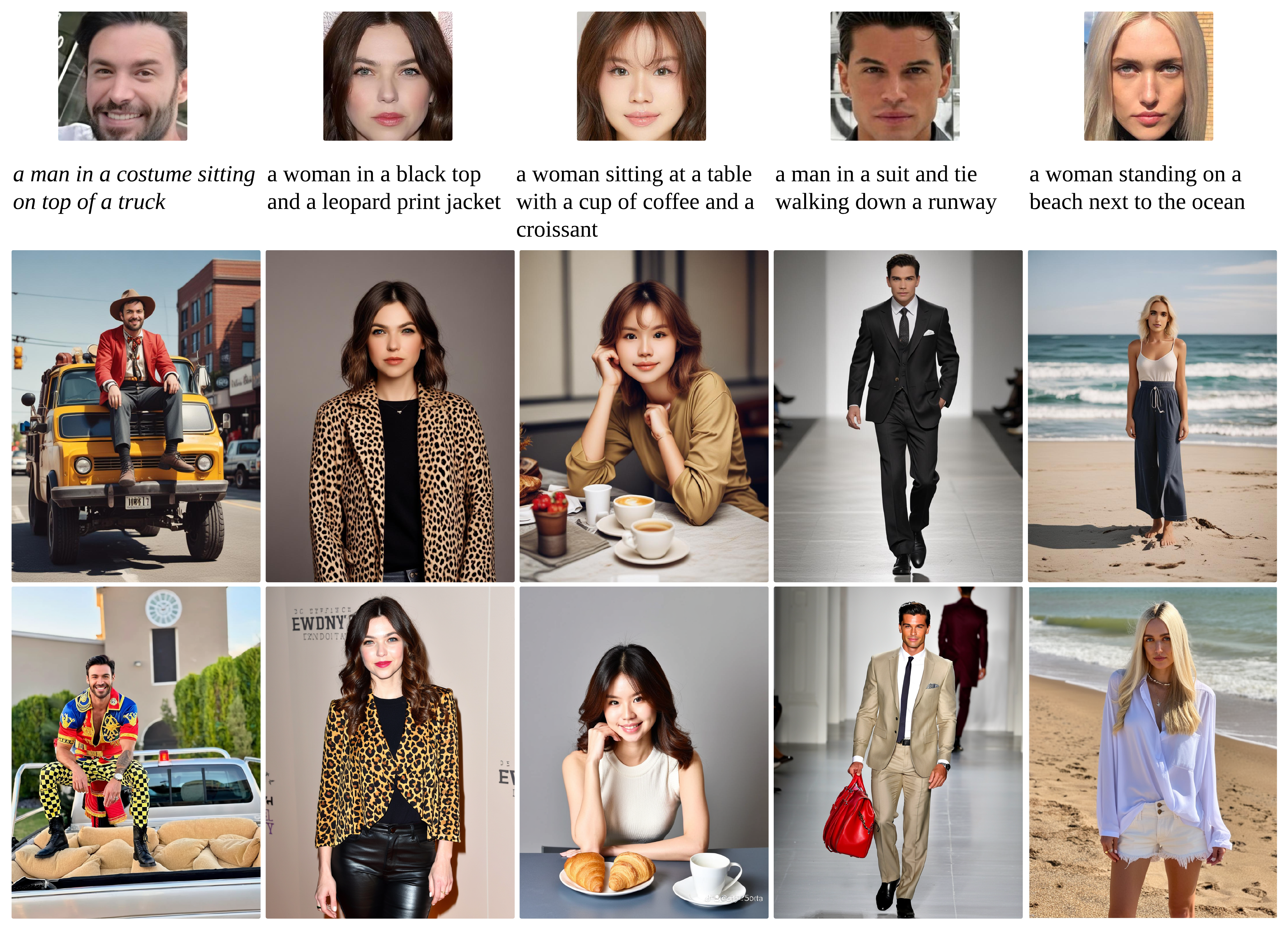}
% \vspace{-0.1in}
\caption{Images produced by FaceCLIP-SDXL and PuLID-SDXL: The first row are the reference images and corresponding text prompts; The second row are the images generated by PuLID-SDXL; The third row are the images generated by FaceCLIP-SDXL.}
\label{fig:comparison}
% \vspace{-0.4cm}
% \vspace{-0.1in}
\end{figure}

\noindent
\textbf{Quantitative Assessment.} We compare FaceCLIP-SDXL against existing ID-preserving image synthesis methods in terms of identity similarity, text adherence, and overall image fidelity. As reported in Table~\ref{tab:comparison}, FaceCLIP-SDXL achieves state-of-the-art performance on the Internal-v1 dataset, with a face similarity score of 0.869 and an FID of 86.9. Additionally, its CLIP score slightly surpasses those of PuLID and InstantID. On the Unsplash-50 dataset, FaceCLIP-SDXL achieves a face similarity score of 0.866 and an FID of 87.8. Qualitative comparisons are shown in Figure~\ref{fig:comparison}. Images generated by PuLID exhibit a strong CG-like appearance and unnatural lighting effects. Moreover, semantic elements are sometimes missing—for example, the croissant in the fifth column is absent. In contrast, FaceCLIP-SDXL produces highly photorealistic images with better identity preservation and precise adherence to text semantics.
\begin{table*}[h]
\small
\centering
\begin{tabularx}{0.9\textwidth}{@{}c|ccc|ccc@{}}
\toprule
\small
\multirow{2}{*}{Model} & \multicolumn{3}{c}{Internal-v1} & \multicolumn{3}{c}{Unsplash-50} \\ 
& Face Sim. $\uparrow$ & CLIP-T $\uparrow$  & FID $\downarrow$ & Face Sim. $\uparrow$ & CLIP-T $\uparrow$  & FID $\downarrow$ \\
\midrule
InstantID & 0.658 &0.302 & 100.6 & 0.664 & 0.309 & 98.5 \\ 
PuLID-SDXL & 0.651 & 0.319 & 112.4 & 0.659 & 0.317 & 112.5 \\
FaceCLIP-SDXL$^{\spadesuit}$ & \textbf{0.869} & \textbf{0.332} & \textbf{86.9} & \textbf{0.866} & \textbf{0.329} & \textbf{87.8} \\
\bottomrule
\end{tabularx}
% \vspace{-0.2cm}
\caption{Peer comparison of FaceCLIP-SDXL, InstantID, and PuLID-SDXL.}
\label{tab:comparison}
% \vspace{-0.4cm}
\end{table*}

\noindent
\textbf{User Study.} We conducted a user study to compare FaceCLIP-SDXL and PuLID-SDXL in terms of identity similarity, text alignment, and image quality. FaceCLIP-SDXL was preferred in 68.6\% of cases, while PuLID-SDXL received 27.8\% of the votes; 8.6\% of responses indicated no preference. These results further confirm that FaceCLIP-SDXL is more favorable from a human perception standpoint.

\noindent
% \subsection{Ablation Study.} To evaluate the impact of our pre-training methodology, we conduct an ablation study by comparing two training variants of FaceCLIP-SDXL. We first pre-train the FaceCLIP-L-14 and FaceCLIP-bigG-14 encoders, and then apply them to train two FaceCLIP-SDXL variants. The first variant, FaceCLIP-SDXL$^{\clubsuit}$, is trained by fine-tuning both the UNet and FaceCLIP’s fusion module. The second variant, FaceCLIP-SDXL$^{\spadesuit}$, is trained by fine-tuning only the UNet, keeping the fusion module fixed. As shown in Table~\ref{tab:ablation}, fine-tuning the fusion module improves text alignment but slightly reduces face similarity. Image quality, measured by FID, remains largely unchanged across the two configurations. These results indicate that, under the encoder-plus-base-model architecture, training solely with diffusion loss tends to bias the model toward generic T2I synthesis objectives. This highlights the critical role of our pre-training strategy in enabling effective identity preservation.
\subsection{Ablation Study} In this section, we investigate the impact of the ID alignment and text alignment losses—$\mathcal{L}_c(e_{c \to r}, e_{r_{\text{cls}}})$ and $\mathcal{L}_c(e_{c \to t}, e_{t_{\text{cls}}})$, respectively—on the model's ability to preserve identity, align with text prompts, and maintain image quality. We train three variants of the FaceCLIP encoder using different loss configurations, as indicated in the first column of Table~\ref{tab:ablation}. $\mathcal{L}_1$ corresponds to the original text-image alignment loss proposed in CLIP~\cite{radford2021learning}, which imparts basic text-to-image generation capability to the model. $\mathcal{L}_2$ and $\mathcal{L}_3$ further incorporate alignment with the facial embedding space and text embedding space, respectively. As shown in Table~\ref{tab:ablation}, the FaceCLIP encoder trained solely with $\mathcal{L}_1$ (CLIP loss) fails to provide identity preservation in the generative model. In contrast, the encoder trained with $\mathcal{L}_2$, which combines CLIP loss and ID alignment loss, enables effective identity preservation. Furthermore, as demonstrated in the third row of Table~\ref{tab:ablation}, incorporating the text alignment loss $\mathcal{L}_c(e_{c \to t}, e_{t_{\text{cls}}})$ introduces beneficial regularization, enhancing the model's text alignment performance.

% We verify the effectiveness of the proposed pre-training loss by progressively adding ID alignment term $\mathcal{L}_c(e_{c \to r}, e_{r_{\text{cls}}})$ and text alignment term $\mathcal{L}_c(e_{c \to t}, e_{t_{\text{cls}}})$ to image-text alignment terms $\mathcal{L}_c(e_{c \to t}, e_I)$. Notably, $\mathcal{L}_c(e_{c \to t}, e_I)$ is exactly the loss proposed in \cite{radford2021learning} to train CLIP image encoder and text decoder.

\begin{table*}
\small
\centering
\begin{tabularx}{0.9\textwidth}{@{}l|ccc@{}}
\toprule
Pre-training Loss &  Face Sim. $\uparrow$ & CLIP-T $\uparrow$  & FID $\downarrow$ \\ 
\midrule
$\mathcal{L}_1 = \mathcal{L}_c(e_{c \to t}, e_I)$ & 0.077 & \textbf{0.351} & 86.9 \\
$\mathcal{L}_2 = \mathcal{L}_c(e_{c \to t}, e_I) + \mathcal{L}_c(e_{c \to r}, e_{r_{\text{cls}}})$ & 0.869 & 0.316 & 86.9\\
$\mathcal{L}_3 = \mathcal{L}_c(e_{c \to t}, e_I) + \mathcal{L}_c(e_{c \to r}, e_{r_{\text{cls}}}) + \mathcal{L}_c(e_{c \to t}, e_{t_{\text{cls}}})$ & \textbf{0.869} & 0.332 & \textbf{86.9}  \\
\bottomrule
\end{tabularx}
% \vspace{-0.05in}
\caption{Impact of different pre-training losses on the final generation quality. The first column lists the loss functions used during the pre-training stage, while the remaining columns report the performance of FaceCLIP-SDXL on various visual metrics.}
\label{tab:ablation}
\vspace{-0.2in}
\end{table*}

\section{Conclusion}
\vspace{-0.05in}
\noindent
In this paper, we introduced FaceCLIP, a novel multimodal encoder that effectively balances subject identity preservation and textual adherence in ID-preserving image synthesis. By learning a joint ID-text representation through multimodal alignment, FaceCLIP captures both identity-specific and semantic attributes, addressing the limitations of existing tuning-based and tuning-free methods. We integrated FaceCLIP with Stable Diffusion XL to create FaceCLIP-SDXL, a state-of-the-art ID-preserving image generation pipeline. Our experimental results demonstrate that FaceCLIP-SDXL outperforms prior identity-preserving models in terms of ID similarity, text adherence, and image quality, achieving new benchmarks on Internal-v1 and Unsplash-50 datasets. Our pre-training approach significantly enhances the encoder’s ability to preserve identity while ensuring semantic flexibility, which is critical for high-fidelity, editable portrait synthesis. By bridging the gap between identity preservation and textual flexibility, FaceCLIP provides a scalable and efficient solution for ID-preserving image generation. 

% \clearpage
{
\bibliographystyle{ieee_fullname}
\bibliography{main}
}

\end{document}